\documentclass[letterpaper]{article} % DO NOT CHANGE THIS
\usepackage[]{aaai25}  % DO NOT CHANGE THIS
\usepackage{times}  % DO NOT CHANGE THIS
\usepackage{helvet}  % DO NOT CHANGE THIS
\usepackage{courier}  % DO NOT CHANGE THIS
\usepackage[hyphens]{url}  % DO NOT CHANGE THIS
\usepackage{graphicx} % DO NOT CHANGE THIS
\usepackage{appendix}
\usepackage{natbib}  % DO NOT CHANGE THIS AND DO NOT ADD ANY OPTIONS TO IT
\usepackage{caption} % DO NOT CHANGE THIS AND DO NOT ADD ANY OPTIONS TO IT
\usepackage{algorithm}
\usepackage{algorithmic}
\usepackage{multirow}
\usepackage{subfigure}
\usepackage{amssymb}
\usepackage{amsmath}
\usepackage{xcolor} 
\usepackage{newfloat}
\usepackage{listings}
\DeclareCaptionStyle{ruled}{labelfont=normalfont,labelsep=colon,strut=off} % DO NOT CHANGE THIS
\floatstyle{ruled}
\newfloat{listing}{tb}{lst}{}
\floatname{listing}{Listing}
\title{Highly Efficient Rotation-Invariant Spectral Embedding for Scalable Incomplete Multi-View Clustering}
\author{
    Xinxin Wang\textsuperscript{\rm 1},  Yongshan Zhang\textsuperscript{\rm 2},  Yicong Zhou\textsuperscript{\rm 1}\thanks{Corresponding author.}\\
}
\affiliations{
    \textsuperscript{\rm 1}Department of Computer and
Information Science, University of Macau\\
    \textsuperscript{\rm 2}School of Computer Science, China University of Geosciences\\ 
    xinxinwang1024@gmail.com, yszhang.cug@gmail.com, yicongzhou@um.edu.mo
}

\usepackage{bibentry}
\begin{document}

\maketitle

\begin{abstract}
Incomplete multi-view clustering presents significant challenges due to missing views. Although many existing graph-based methods aim to recover missing instances or complete similarity matrices with promising results, they still face several limitations: (1) Recovered data may be unsuitable for spectral clustering, as these methods often ignore guidance from spectral analysis; (2) Complex optimization processes require high computational burden, hindering scalability to large-scale problems; (3) Most methods do not address the rotational mismatch problem in spectral embeddings. To address these issues, we propose a highly efficient rotation-invariant spectral embedding (RISE) method for scalable incomplete multi-view clustering. RISE learns view-specific embeddings from incomplete bipartite graphs to capture the complementary information. Meanwhile, a complete consensus representation with second-order rotation-invariant property is recovered from these incomplete embeddings in a unified model. Moreover, we design a fast alternating optimization algorithm with linear complexity and promising convergence to solve the proposed formulation. Extensive experiments on multiple datasets demonstrate the effectiveness, scalability, and efficiency of RISE compared to the state-of-the-art methods. 
\end{abstract}

%     \link{Code}{https://github.com/RISE2025}
%     % \link{Datasets}{https://aaai.org/example/datasets}
%     % \link{Extended version}{https://aaai.org/example/extended-version}

\section{Introduction}
Multi-view clustering (MVC) explores the affinity relationships of samples among multiple views and groups similar samples into the same cluster \cite{ zhangchangqing2018generalized}. Over the past decade, a variety of MVC methods have been proposed for various real-world applications \cite{chenmansheng2020multi, li2024tensorized}. Although these methods achieve superior clustering performance, most of them strictly rely on the assumption that all instances in each view are available and complete. However, in real-world scenarios, some instances may be inaccessible from certain views due to transmission loss or sensor malfunctions \cite{wenjie2020adaptive}. For instance, in medical diagnosis, patients may undergo various pathology tests, such as MRI, CT, and genetic testing. Only a small percentage of patients require all these tests, while others may need only a few targeted assessments. The resulting multi-view data, which contains missing instances, is referred to as incomplete multi-view data. When existing MVC methods are applied to such incomplete datasets, they may experience significant performance degradation or even complete failure \cite{xiawei2022tensor, shenqiangqiang2023robust}.

\begin{figure}[t]
  \centering
    % \fbox{\rule{0pt}{2.5in} \rule{0.9\linewidth}{0pt}}
  \includegraphics[width=\linewidth]{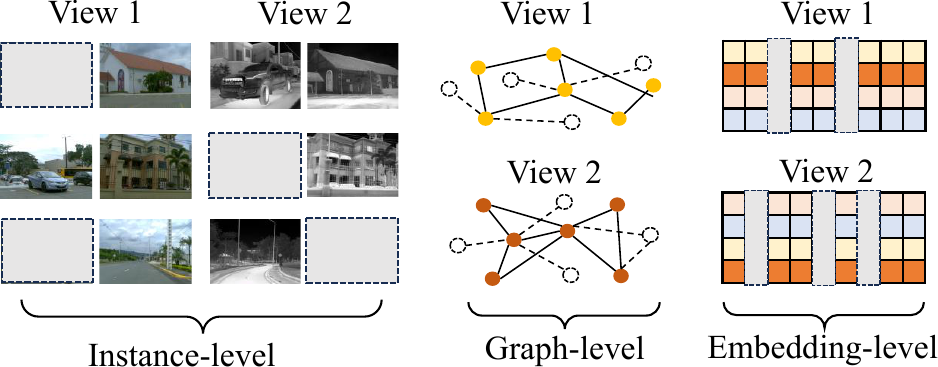}
  \caption{Different completion stages for IMVC. Filling missing data at the feature embedding level is computationally efficient and provides robust performance.}
\label{fig-incompleteScheme}
\vspace{-10pt}
\end{figure}

To solve this problem, incomplete multi-view clustering (IMVC) has gained increasing attention. Recently, a variety of IMVC methods have been proposed  \cite{wen2020incomplete, wenjie2021generalized, lizhenglai2022high}. These methods can be divided into three categories: instance-level, graph-level, and embedding-level methods. As shown in Fig. \ref{fig-incompleteScheme} (a), instance-level methods are natural and intuitive, as they recover missing instances to facilitate subsequent clustering \cite{ haowenyu2023tensor, zhaoshuping2023tensorized, liujiyuan2021self}. 
Typical recovery strategies include simple averaging \cite{shao2015multiple, gaohang2016incomplete}, matrix factorization \cite{yinjun2021incomplete, wenjie2019unified}, and deep learning-based imputation \cite{linyijie2021completer}.
%Recovery strategies such as simple averaging, matrix factorization, and deep learning-based completion may lead to noise sensitivity or involve complex optimization processes. 
As shown in Fig. \ref{fig-incompleteScheme} (b), graph-level methods aim to fill in incomplete similarity matrices rather than directly recovering missing instances \cite{wenjie2023highly, cuijinrong2022low, liucheng2023self, zhangchao2023enhanced, lvziyu2022view, zhangguangyu2024unified}. During the learning process, the recovered similarity matrices can either be used in spectral clustering or employed to directly output clustering results based on the connected components.
%Although remarkable success has been achieved,  graph completion always requires at least $\mathcal{O}(n^2)$ complexity in both computation and storage, making it impractical for large-scale problems. 
Unlike instance-level and graph-level completion methods, as shown in Fig. \ref{fig-incompleteScheme}(c), embedding-level methods perform completion in the latent feature space \cite{zhangchao2023robust}. This approach efficiently recovers missing data from low-dimensional embeddings, and the recovered representations naturally align with clustering tasks.

\begin{figure}[t]
  \centering
    % \fbox{\rule{0pt}{2.5in} \rule{0.9\linewidth}{0pt}}
  \includegraphics[width=0.9\linewidth]{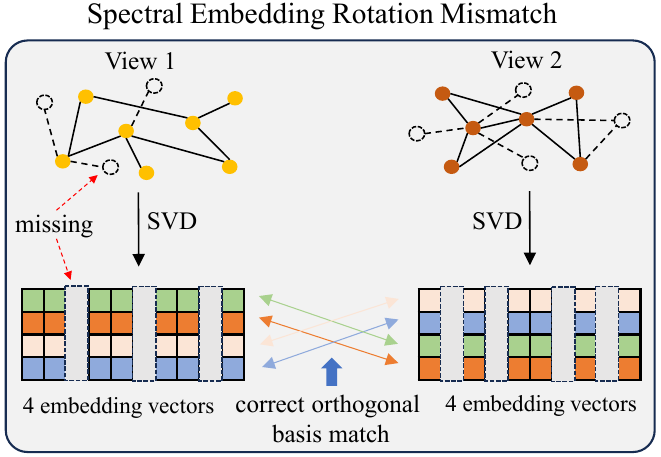}
  \caption{Illustration of the SERM problem. Each color row represents a relaxed orthogonal basis. With structure drift in Laplacian matrices for incomplete views, the resulting orthogonal spectral embedding may be rotational mismatch.}
 \label{SERM}
 \vspace{-10pt}
\end{figure}

Although there are viable precedents for filling in missing data at different completion stages, existing methods still exhibit several limitations. \textit{First}, most instance-level and graph-level methods finish multi-view fusion and completion before spectral analysis, which may render the recovered data unsuitable for spectral clustering \cite{zhangchao2023robust}. \textit{Second}, most graph-level methods involve complex optimization processes and require a computational burden of at least $\mathcal{O}(n^2)$ in both computation and storage, making them impractical for large-scale problems \cite{liusuyuan2022efficient, wangsiwei2022highly}. \textit{Third}, as is shown in Fig. \ref{SERM}, disparities in available instances among views cause clustering structure shifts across similarity graphs and increase the imbalance of eigenvalue distributions among the Laplacian matrices. This may worsen potential feature rotational mismatches between relaxed spectral embedding matrices when performing SVD decomposition. We refer to this issue as the spectral embedding rotational mismatch (SERM) problem. A similar issue in multi-view subspace learning is referred to as the misalignment problem \cite{wenyi2023scalable, wangsiwei2022align}. To the best of our knowledge, no unified IMVC framework simultaneously addresses these issues.

To address these issues, this paper proposes a highly efficient rotation-invariant spectral embedding (RISE) method for scalable incomplete multi-view clustering. Specifically, we learn view-specific embeddings from incomplete bipartite graphs to capture discrepancies between views. To tackle the SERM problem among view-specific embeddings, we propose a second-order rotation-invariant learning module to recover the complete consensus representation. Both processes are integrated into a unified framework, where the view-specific embedding can better serve for the learning of complete consensus representation, and the latter is able to guide the learning of the former. To solve the proposed formulation, we propose a fast alternating optimization algorithm with promising convergence, reducing the time complexity from $\mathcal{O}(n^3v)$ to $\mathcal{O}(nmv)$. The storage cost of RISE is approximately $\mathcal{O}(n)$ by using bipartite graphs. The main contributions of this paper are listed below. 

\begin{itemize}
  \item  We propose an efficient IMVC method, termed as RISE. It integrates incomplete view-specific embedding learning and complete consensus representation learning in a unified model. By introducing incomplete bipartite graph, RISE shows scalability and high-efficiency. 
  \item To address the SERM problem, we propose a second-order rotation-invariant learning module to recover the complete consensus representation from view-specific embeddings. Guided by this module, the cross-view spectral embeddings can achieve adequate consistency. 
  \item We design a fast two-step optimization algorithm for RISE with theoretical analyses. Our proposed algorithm demonstrates promising convergence and linear time complexity with respective to the number of samples. Comprehensive experiments validate the effectiveness and efficiency of our method.
\end{itemize}

\section{Related Work}
To tackle the scalability and efficiency issues in IMVC, several bipartite graph-based approaches have achieved significant attention. Wang \textit{et al.} \cite{wangsiwei2022highly} first introduced the bipartite graph for IMVC. Li \textit{et al.} \cite{li2024parameter} proposed a scalable parameter-free model, while Liu \textit{et al.} \cite{liusuyuan2022fast} developed a fast model with view-independent anchors. Zhao \textit{et al.} \cite{zhao2023self} refined the complete bipartite graph learning by using concatenated multiple features. These methods learn a complete unified bipartite graph across views, to which spectral clustering is applied to obtain final clustering results. However, enforcing consistent cross-view representation before spectral analysis can neglect inter-view discrepancies, leading to suboptimal spectral embedding representations. Recent works have integrated tensor learning techniques to explore the inter-view low-rank structure. Chen \textit{et al.} \cite{chen2023tensor} combined tensor learning with view-specific bipartite graphs, while Li \textit{et al.} \cite{li2023distribution} integrated the consistent distribution between anchors and observed embedded representations to improve anchor imputation. Long \textit{et al.} \cite{long2024feature} applied tensor networks to explore both cross-view and structural correlations. However, computing high-order correlations across views using tensor singular value decomposition (t-SVD) or tensor ring (TR) decomposition can be time-consuming, which limits their applicability in real-world scenarios. 

\section{Proposed Method}
\subsection{Formulation}
Recently, several works have successfully adopted bipartite graphs to enhance the model's scalability and clustering efficiency in IMVC tasks \cite{wangsiwei2022highly,li2024parameter}.
Similar to previous works \cite{hefang2020fast}, we use bipartite graph to reduce structural redundancy when presenting the correlations among samples. Specifically, given incomplete multi-view data $\left\{X^{(i)} \in \mathbb{R}^{d_i \times n_i}\right\}_{i=1}^v$ with $d_i$ features and $n_i$ samples in $v$ views, $K$-means is used to generate $m$ anchors $A^{(i)}=[a_1,a_2,...,a_m]\in \mathbb{R}^{d_i \times m}$  for each view separately. Then, a bipartite graph $\overline{B}^{(i)} \in \mathbb{R}^{n_i \times m}$ is constructed between $m$ anchors and $n_i$ samples on each view individually, where each element in $\overline{B}^{(i)}$ is computed by
\begin{equation}
\label{eq:2-4}
b_{p,q}= \begin{cases}\frac{d(p, k+1)-d(p, q)}{k d(p, k+1)-\sum_{q=1}^k d(p, q)} & \forall q \in \Phi_p \\ \quad  0 & \forall q \notin \Phi_p\end{cases},
\end{equation}
where $\Phi_p$ is the index set of the $k$ nearest anchors of sample $x_p$, and $d(p, q)=\left\|\mathbf{x}_p-\mathbf{a}_q\right\|^2 $ presents the Euclidean distance between $x_p$ and its $q$-th nearest anchor $a_q$. 

Rather than using full Laplacian matrices, we define a highly efficient spectral embedding learning framework for scalable incomplete multi-view clustering based on bipartite graphs as follows:
\begin{equation}
\label{eq:3-1}
% \resizebox{1\hsize}{!}{$
  \max _{F^(i)}\sum_{i=1}^v \operatorname{Tr}\left(F^{(i)^T} B^{(i)} B^{(i)^T} F^{(i)}\right), \text{~s.t.~}  F^{(i)^T}F^{(i)}=I_{k},
\end{equation}
where $B^{(i)}=\overline{B}^{(i)}\Lambda^{(i)-1/2}$ is the normalized bipartite graph of the $i$-th view, and $F^{(i)} \in \mathbb{R}^{n_i \times k}$ is the corresponding embedding generated from $B^{(i)}$. It should be noted that $\Lambda^{(i)} \in \mathbb{R}^{m \times m}$ is a diagonal matrix with diagonal elements given by $\Lambda^{(i)}_{q,q} = \sum_{p=1}^{n_i} \overline{B}^{(i)}_{p,q}$. 

By solving Eq. \eqref{eq:3-1}, we obtain the embedding $F^{(i)} \in \mathbb{R}^{n_i \times k}$ with varying numbers of available feature representations. Afterward, a question that naturally arises is: \textbf{How to efficiently and effectively refine the cross-view spectral embedding match and recover the consensus complete embedding?} An common solution is to adopt the following formula to learn the complete consensus representation from incomplete view-specific embeddings \cite{wenjie2021generalized} by
\begin{equation}
\label{eq:3-1-1}
% \resizebox{0.5\hsize}{!}{$
 \min _{Y}\sum_{i=1}^v\left\|F^{(i)}-Q^{(i)^T} Y\right\|_F^2,
\end{equation}
where $Q^{(i)} \in \{0,1\}^{n \times n_i}$ is an index matrix of the $i$-th view that is used to remove the entries corresponding to the unavailable instances from the complete consensus representation $Y \in  \mathbb{R}^{n \times k}$.  $Q^{(i)}$ is constructed from the index vector $h^{(i)} \in \mathbb{R}^{n_i}$, which records the indices of the $n_i$ instances in $F^{(i)}$ within the complete dataset. For the $q$-th column of $Q^{(i)}$, the elements are as follows: 
\begin{equation}
\label{eq:3-3}
% \resizebox{0.7\hsize}{!}{$
\begin{aligned}
  Q_{p q}^{(i)}= \begin{cases}1, & \text { if } p \text { equals to the } q \text {-th}\text { index in } h^{(i)} \\0, & \text { otherwise. }\end{cases}
\end{aligned}
\end{equation}
However, the recovery strategy in Eq. \eqref{eq:3-1-1} only considers first-order correlations during information fusion and overlooks the potential SERM problem. It may damage the representation capability of the model. Inspired by \cite{xiadongxue2023Partion}, we address this issue by recovering the complete second-order correlations among the embeddings of all samples in a quadratic term form. 

The separate processes of spectral embedding learning and complete consensus representation learning may result in suboptimal results. Therefore, we combine them into a unified formula as follows:
\begin{equation}
\label{eq:3-2}
\resizebox{0.7\hsize}{!}{$
\begin{aligned}
  \min_{Y,F^{(i)}} & \sum_{i=1}^v \left\|YY^T-Q^{(i)}F^{(i)}F^{(i)^T}Q^{(i)^T}\right\|_F^2 \\ & - \beta \sum_{v=1}^v \operatorname{Tr}\left(F^{(i)^T} B^{(i)} B^{(i)^T} F^{(i)}\right), \\
  &\quad \text{s.t.} \quad  F^{(i)^T}F^{(i)} = I, \quad  Y^TY = I,  
\end{aligned} $}
\end{equation}
where the first term indicates the complete consensus representation learning with second-order rotation-invariant property, and the second term represents view-specific spectral embedding learning. Specifically, we impose an orthogonality constraint on $Y$ to enhance its discriminative power. $\beta$ is the penalty parameter, and $Q^{(i)}F^{(i)}$ denotes the recovery of the original arrangement.  
It is easy to prove that the formulation in Eq. \eqref{eq:3-2} is rotation-invariant \cite{von2007tutorial}. This implies that for a spectral embedding $F^{(i)}_*$, given any orthogonal rotation matrix $R$, $F^{(i)}_* R$ is a feasible spectral embedding set for Eq. \eqref{eq:3-2}. Thus, RISE is unaffected by potential rotations and always obtains the optimal solution $\hat{Y}$. 

Although bipartite graphs can reduce space complexity, simply introducing the index matrix $Q^{(i)} \in \mathbb{R}^{n \times n_i}$ incurs a space burden of $\mathcal{O}(n^2)$ and computational overhead of $\mathcal{O}(n^3)$, which cam become a serious issue for large-scale data. To address this problem, we innovatively define two operations for large-scale IMVC issues: $Q^{(i)}F^{(i)}=\zeta^{-1}(F^{(i)}, h^{(i)})$, $Q^{(i)^T}Y=\zeta(Y, h^{(i)})$, where $\zeta(\cdot)$ and $\zeta^{-1}(\cdot)$ denote the selection of instances from the complete representation matrix according to the index vector and its inverse operation, respectively. With this transformation, the space complexity drops from $\mathcal{O}(vn^2)$ to $\mathcal{O}(vn)$.

\subsection{Optimization}
%The optimization problem of Eq. \eqref{eq:3-2} is not jointly convex to all variables simultaneously. 
To solve the optimization problem in Eq. \eqref{eq:3-2}, we devise a fast alternating optimization algorithm to update each variable while keeping the other variables fixed.

\textbf{Updating} $F^{(i)}$: When $Y$ is fixed, the optimization problem \textit{w.r.t} $F^{(i)}$ is
% \begin{equation}
% \label{eq:3-4}
% \resizebox{1\hsize}{!}{
% $\begin{aligned}
%   \max_{ F^{(i)^T}F^{(i)} = I} & \operatorname{Tr} \left(2F^{(i)^T} Q^{(i)^T}YY^TQ^{(i)} F^{(i)} \right) + \beta  \operatorname{Tr}\left(F^{(i)^T} B^{(i)} B^{(i)^T} F^{(i)}\right)
% \end{aligned}$}
% \end{equation}
\begin{equation}
\label{eq:3-4}
\resizebox{1\hsize}{!}{$
\begin{aligned}
  \max_{ F^{(i)^T}F^{(i)} = I} & \operatorname{Tr} \left(2F^{(i)^T} \hat{Y}\hat{Y}^T F^{(i)} \right) + \beta  \operatorname{Tr}\left(F^{(i)^T} B^{(i)} B^{(i)^T} F^{(i)}\right),
\end{aligned}$}
\end{equation}
where $\hat{Y}=Q^{(i)^T}Y$. The solution to Eq. \eqref{eq:3-4} is the $k$ eigenvectors corresponding to the largest $k$ eigenvalues of  $S^{(i)}=2\hat{Y}\hat{Y}^T+\beta B^{(i)}B^{(i)^T}$. However, performing eigen decomposition requires at least $\mathcal{O}(n_i^2k)$ time complexity, which makes it impractical for large-scale problems.

In this work, we present an alternative approach to efficiently compute the $k$ eigenvectors of matrix $S^{(i)}$ based on proposition 1. 

\textbf{\textit{Proposition 1.}} Given a set of similarity matrices $\left\{S_t\right\}_{t=1}^\pi \in \mathbb{R}^{n \times n} $, where each matrix satisfies $S_t = Z_tZ_t^T$. Defining the concatenated matrix $Z_{cat} = [Z_1,...,Z_t,...,Z_\pi] \in \mathbb{R}^{n \times (m_1+m_2+...m_\pi)}$ and supposing the singular value decomposition (SVD) of $Z_{cat}$ is $U \Sigma V^T$ (with $UU^T = I, V^TV = I$), we have that:
\begin{equation}
\label{eq:3-5}
% \resizebox{0.8\hsize}{!}{$
\begin{aligned}
  \max_{ F^TF = I} \operatorname{Tr} \left(F^T (\sum_{t=1}^\pi S_t) F \right)  \Leftrightarrow \min _{F^T F= I}\left\|Z_{cat}-F F^T Z_{cat}\right\|_F^2 .
\end{aligned} 
\end{equation}
In Eq. \eqref{eq:3-5}, the optimal solution of $F$ is equal to $U$.

\textbf{\textit{Proof}.} Based on the right-hand side of Eq. \eqref{eq:3-5}, we can establish the following equivalences:
\begin{equation}
% \resizebox{0.8\hsize}{!}{$
\begin{aligned}
   \min _{F^T F= I}\left\|Z_{cat}-F F^T Z_{cat}\right\|_F^2  &\Leftrightarrow \max _{F^T F= I}\operatorname{Tr} \left(F^T Z_{cat}Z_{cat}^T F \right)\\
   & \Leftrightarrow  \max_{ F^TF = I} \operatorname{Tr} \left(F^T  (\sum_{t=1}^\pi S_t) F \right) 
\nonumber 
\end{aligned} 
\end{equation}
Moreover, 
\begin{equation}
% \resizebox{0.6\hsize}{!}{$
\begin{aligned}
    \sum_{t=1}^\pi S_t = Z_{cat}Z_{cat}^T = (U \Sigma V^T) (U \Sigma V^T)^T 
    = U \Sigma^2 U^T.
\nonumber 
\end{aligned} 
\end{equation}

To solve Eq. \eqref{eq:3-4}, we let $Z_{cat}^{F(i)} = [ \sqrt{2} Q^{(i)^T}Y,  \sqrt{\beta} B^{(i)}] \in \mathcal{R}^{n_i \times (k+m)}$. According to \textit{Proposition 1}, the optimal solution to Eq. \eqref{eq:3-4} is the $k$ left singular vectors of $Z_{cat}^{F(i)}$. This approach naturally reduces the computational cost of updating $F^{(i)}$ from $\mathcal{O}(n_i^2k)$ to $\mathcal{O}(n_i(k+m)^2)$.

\textbf{Updating} $Y$: When $F^{(i)}$ is fixed, the optimization problem \textit{w.r.t} $Y$ is
\begin{equation}
\label{eq:3-6}
% \resizebox{0.7\hsize}{!}{$
\begin{aligned}
  \max_{Y^TY = I} & \operatorname{Tr} \left(Y^T \left(\sum_{i=1}^v Q^{(i)}F^{(i)}F^{(i)^T}Q^{(i)^T} \right) Y \right). 
\end{aligned}
\end{equation}
Based on \textit{Proposition 1}, we define the formulation $Z_{cat}^Y = [Q^{(1)}F^{(1)},...,  Q^{(i)}F^{(i)},..., Q^{(v)}F^{(v)}] \in \mathbb{R}^{n \times (vk)}$. Similarly, the optimal solution to Eq. \eqref{eq:3-6} is the $k$ left singular vectors of $Z_{cat}^Y$. Updating $Y$ requires only $\mathcal{O}(n(vk)^2)$ time complexity.

The entire optimization procedure for RISE is summarized in Algorithm \ref{alg1}. The code is available at https://github.com/RISE2025.

\begin{algorithm}[!t]
    \renewcommand{\algorithmicrequire}{\textbf{Input:}}
	\renewcommand{\algorithmicensure}{\textbf{Initialize:}}
    \caption{RISE}
    \label{alg1}
    %\begin{flushleft}
    %\textbf{Input:} 111\\
    %\textbf{Output:} 222
    %\end{flushleft}
    \begin{algorithmic}[1]
    \REQUIRE Normalized bipartite graph $\{B^{(i)})\}_{i=1}^v$,  available index $\{h^{(i)})\}_{i=1}^v$, number of cluster $c$, and embedding dimension $k$.
    \ENSURE Initialize $F^{(i)}$ by performing $SVD$ on $B^{(i)}$.
    \WHILE {not converge}
        \STATE Update $Y$ by solving Eq. \eqref{eq:3-6};
        \STATE Update $\{F^{(i)}\}_{i=1}^v$ by solving Eq. \eqref{eq:3-4}; 
    \ENDWHILE
        \STATE \textbf{Output:} perform $k$-means on $Y$ to achieve the final clustering indicator.
    \end{algorithmic}
\end{algorithm}

\subsection{Algorithm Analysis}
% Firstly, we analyze the space and time complexities of our algorithm. Then, the convergence of RISE-IMVC is proven theoretically.
\textbf{Space complexity.} The storage burden for RISE are the matrices $B^{(i)} \in \mathbb{R}^{n_i \times m}$, $F^{(i)} \in \mathbb{R}^{n_i \times k}$, and $Y \in \mathbb{R}^{n \times k}$. Thus, the space complexity of  RISE is $\mathcal{O}(nk+n_ik+n_im)$.

\noindent\textbf{Time complexity.} The time complexity of RISE involves updating $F^{(i)}$ and $Y$. Updating $F^{(i)}$ takes $\mathcal{O}(n_i(k+m)k)$, while and solving for $Y$ costs $\mathcal{O}(nvk^2)$. Thus, at each iteration, the time complexity is $\mathcal{O}( \sum_{i=1}^v n_i(k+m)k + nvk^2 )$. After optimization, we perform $k$-means with $\mathcal{O}(nc)$ to output the discrete clustering indicators. Considering that $ k \ll n $, $ m \ll n $, and $ c \ll n $, the overall time complexity of the proposed optimization algorithm is linear to the number of samples $n$. This demonstrates the high efficiency of RISE. 

\begin{table}[!t]
\scriptsize
\begin{tabular}{ccccc}
\hline
Dataset                         & Size   & Views & Classes & Dimensions             \\ \hline
\multicolumn{1}{l}{Prokaryotic} & 551    & 3     & 4       & 438,3,393              \\
WebKB                           & 1051   & 2     & 2       & 2949,334               \\
Caltech101-7                    & 1474   & 6     & 7       & 48,40,254,1984,512,928 \\
Wikipedia                       & 2866   & 2     & 10      & 128,10                 \\
CIFAR10                         & 50000  & 3     & 10      & 512,2048,1024          \\
FMNIST                          & 60000  & 3     & 10      & 512,512,1280           \\
YoutubeFace                     & 101499 & 5     & 31      & 64,512,64,647,838      \\ \hline
\end{tabular}
\caption{Descriptions of Multi-view Datasets}
\label{tag:datasets}
\end{table}

\textbf{Convergence.} The optimization algorithm \ref{alg1} updates $F^{(i)}$ and $Y$ by alternatively solving Eq. \eqref{eq:3-4} and Eq. \eqref{eq:3-6}. The SVD-based solution designed in \textit{Proposition 1} converges to the global minimum for each sub-problem. Moreover, the objective function in Eq. \eqref{eq:3-2} has a lower bound. These factors collectively ensure the convergence of the algorithm. More details in the appendix. More details in the appendix.

\section{Experiments}
% In this section, we first introduce our experimental setup in detail, including benchmark datasets and incomplete data construction rule, and compared methods and evaluation metrics. Then we demonstrate extensive experimental results and analysis. 

\subsection{Experimental Settings}
\textbf{Datasets and Incomplete Data Construction.}
We evaluate our RISE method on seven popular multi-view datasets, which include Prokaryotic \cite{brbic2018multi}, Wikipedia \cite{josecp2014role}, WebKB \cite{bisson2012}, 
Caltech101-7 \cite{fei2006one},
CIFAR10 \cite{krizhevsky2009learning}, FMNIST \cite{xiao2017fashion}, and YoutubeFace \cite{YoutubeFace2011}. The details of these datasets are described in Table \ref{tag:datasets}.
% We evaluate our method on seven popular multi-view datasets, consisting of Prokaryotic\footnote{https://github.com/mbrbic/Multi-view-LRSSC/tree/master/datasets}, WikipediaArticles\footnote{http://www.svcl.ucsd.edu/projects/crossmodal/}, WebKB\footnote{http://lig-membres.imag.fr/grimal/data.html}, 
% Caltech101-7\footnote{http://www.vision.caltech.edu/Image Datasets/Caltech101/},
% CIFAR10\footnote{https://www.cs.toronto.edu/kriz/cifar.html}, FMNIST\footnote{http://yann.lecun.com/exdb/mnist/} and YoutubeFace\footnote{https://www.cs.tau.ac.il/ wolf/ytfaces/}. The detail of them are described in Table \ref{tag:datasets}. 
% To be specific, prokaryotic is a prokaryotic species database, including text and different genomic representations. WebKB is a collection of web context and links. WikipediaArticles contains 693 multimedia documents. 
% Caltech101-7 is an image dataset with 1474 instances. 
% Cifar10 is a tiny color image database. fmnist contains 60000 handwritten numerals from 0 to 9. YoutubeFace is a video dataset. 

Following the approaches in \cite{wangsiwei2022highly, wenyi2023scalable}, we construct the incomplete data with 9 different missing rates $p=[0.1:0.1:0.9]$. For example, when $p=0.9$, we randomly select 10\% samples as complete data and randomly drop partial views of the rest 90\% samples.
% To generate incomplete multi-view datasets, we randomly select $n_d$ paired samples are incomplete. Specifically, we generate a random matrix $R = [r_1,r_2,...,r_{n_d}]^T \in \{0,1\}^{n_d \times v}$, ensuring $0<\sum_{i=1}^v r_{l i}<v$. The $l$-th sample is observed in the $h$-th view with $r_{lh}=1$. $r_{lh}=0$ denotes the $l$-th sample in the $h$-th view is missing. For the rest $n-n_d$ samples, we set them to be available in all views. By varying the size of $n_d$, we can manually set the missing ratio $\mu = \frac{n_d}{n}$ in the range of $[0.1:0.1:0.9]$. 
\begin{table*}[!t]
\centering
% \footnotesize
\scriptsize
\renewcommand\arraystretch{0.9}
\tabcolsep = 3.6 pt
\begin{tabular}{ccccccccccccc}
\hline
Metrices                    & Methods                          & UEAF                      & PIMVC       & HCLS-CGL       & GSRIMC      & IMVTSC-MVI & DAIMC                     & sFSR-IMVC   & PSIMVC-PG & IMVC-CBG    & SIMVC-SA       & Ours       \\ \hline
\multicolumn{1}{c|}{}       & Prokaryotic                      & 53.49                     & 54.03       & 56.62          & 49.36       & 53.85      & 51.31                     & 55.13       & 47.16     & 54.91       & {\underline{60.42}}    & \textbf{75.13} \\
\multicolumn{1}{c|}{}       & WebKB                            & 76.56                     & 86.06       & 81.99          & {\underline{87.30}} & 81.54      & 81.77                     & 78.72       & 71.08     & 86.66       & 83.60          & \textbf{93.02} \\
\multicolumn{1}{c|}{}       & \multicolumn{1}{l}{Caltech101-7} & \multicolumn{1}{c}{37.26} & 66.88       & {\underline{72.36}}    & 65.22       & 60.53      & 46.97                     & 67.60       & 48.52     & 59.88       & 59.37          & \textbf{75.25} \\
\multicolumn{1}{c|}{ACC}    & Wikipedia                        & 46.01                     & 46.06       & 39.73          & 47.52       & 45.13      & \multicolumn{1}{c}{46.87} & {\underline{47.98}} & 47.44     & 47.18       & 47.22          & \textbf{48.64} \\
\multicolumn{1}{c|}{}       & CIFAR10                          & -                         & -           & -              & -           & -          & 92.80                     & 62.77       & 95.96     & 95.91       & { \underline{96.04}}    & \textbf{96.54} \\
\multicolumn{1}{c|}{}       & FMNIST                           & -                         & -           & -              & -           & -          & 25.05                     & 10.76       & 22.20     & 22.74       & {\underline{50.99}}    & \textbf{52.49} \\
\multicolumn{1}{c|}{}       & YoutubeFace                      & -                         & -           & -              & -           & -          & \multicolumn{1}{c}{-}     & -           & 14.32     & {\underline{23.81}} & 15.68          & \textbf{25.52} \\ \hline
\multicolumn{1}{c|}{}       & Prokaryotic                      & 27.12                     & 25.87       & 7.28           & 17.02       & 19.54      & 15.98                     & 11.05       & 22.84     & 31.25       & {\underline{32.68}}    & \textbf{38.60} \\
\multicolumn{1}{c|}{}       & WebKB                            & 17.24                     & 23.92       & 17.01          & {\underline{40.91}} & 9.19       & 16.62                     & 2.25        & 10.76     & 34.05       & 25.14          & \textbf{53.61} \\
\multicolumn{1}{c|}{}       & \multicolumn{1}{l}{Caltech101-7} & \multicolumn{1}{c}{27.88} & 54.09       & \textbf{57.51} & 53.18       & 46.52      & 48.09                     & 44.75       & 42.80     & 43.13       & 40.72          & {\underline{54.06}}    \\
\multicolumn{1}{c|}{NMI}    & Wikipedia                        & 36.11                     & 34.11       & 28.27          & 23.75       & 38.94      & \multicolumn{1}{c}{32.37} & 34.36       & 35.78     & {\underline{37.96}} & \textbf{39.18} & 35.78          \\
\multicolumn{1}{c|}{}       & CIFAR10                          & -                         & -           & -              & -           & -          & 89.28                     & 72.79       & 90.56     & 90.60       & {\underline{90.67}}    & \textbf{91.56} \\
\multicolumn{1}{c|}{}       & FMNIST                           & -                         & -           & -              & -           & -          & 10.63                     & 0.06        & 3.11      & 6.01        & {\underline{27.96}}    & \textbf{38.73} \\
\multicolumn{1}{c|}{}       & YoutubeFace                      & -                         & -           & -              & -           & -          & \multicolumn{1}{c}{-}     & -           & {\underline{11.54}}     & 11.51     & 8.58           & \textbf{17.88} \\ \hline
\multicolumn{1}{c|}{}       & Prokaryotic                      & 66.90                     & 63.96       & 59.58          & 61.67       & 58.08      & 58.41                     & 57.75       & 59.04     & 68.81       & {\underline{72.75}}    & \textbf{77.52} \\
\multicolumn{1}{c|}{}       & WebKB                            & 80.90                     & 84.95       & 82.22          & {\underline{88.50}} & 81.54      & 81.65                     & 78.74       & 78.29     & 86.67       & 83.60          & \textbf{93.39} \\
\multicolumn{1}{c|}{}       & \multicolumn{1}{l}{Caltech101-7} & \multicolumn{1}{c}{77.56} & {\underline{86.94}} & 86.66          & 58.03       & 85.48      & 83.33                     & 74.50       & 80.61     & 81.03       & 79.27          & \textbf{88.59} \\
\multicolumn{1}{c|}{Purity} & Wikipedia                        & 49.39                     & 50.77       & 44.48          & 39.10       & 50.76      & \multicolumn{1}{c}{49.02} & 37.17       & 51.10     & 50.89       & {\underline{51.58}}    & \textbf{57.04} \\
\multicolumn{1}{c|}{}       & CIFAR10                          & -                         & -           & -              & -           & -          & 93.15                     & 62.81       & 95.96     & 95.91       & {\underline{96.04}}    & \textbf{96.54} \\
\multicolumn{1}{c|}{}       & FMNIST                           & -                         & -           & -              & -           & -          & 25.55                     & 10.79       & 22.20     & 22.80       & {\underline{51.16}}    & \textbf{60.17} \\
\multicolumn{1}{c|}{}       & YoutubeFace                      & -                         & -           & -              & -           & -          & \multicolumn{1}{c}{-}     & -           & 28.11     & {\underline{28.99}} & 28.08          & \textbf{37.78} \\ \hline
\end{tabular}
\caption{The average ACC, NMI, Purity results of different methods with nine missing ratios on benchmark datasets. ``-'' indicates out of CPU memory. The best results are highlighted in bold, while the second-best results are marked with underline.}
\label{table_performance}
\end{table*}

\begin{figure*}[!t]
\centering
\setlength{\abovecaptionskip}{0.2cm}
\subfigure{
\includegraphics[trim=50 250 60 280, width=0.235\textwidth]{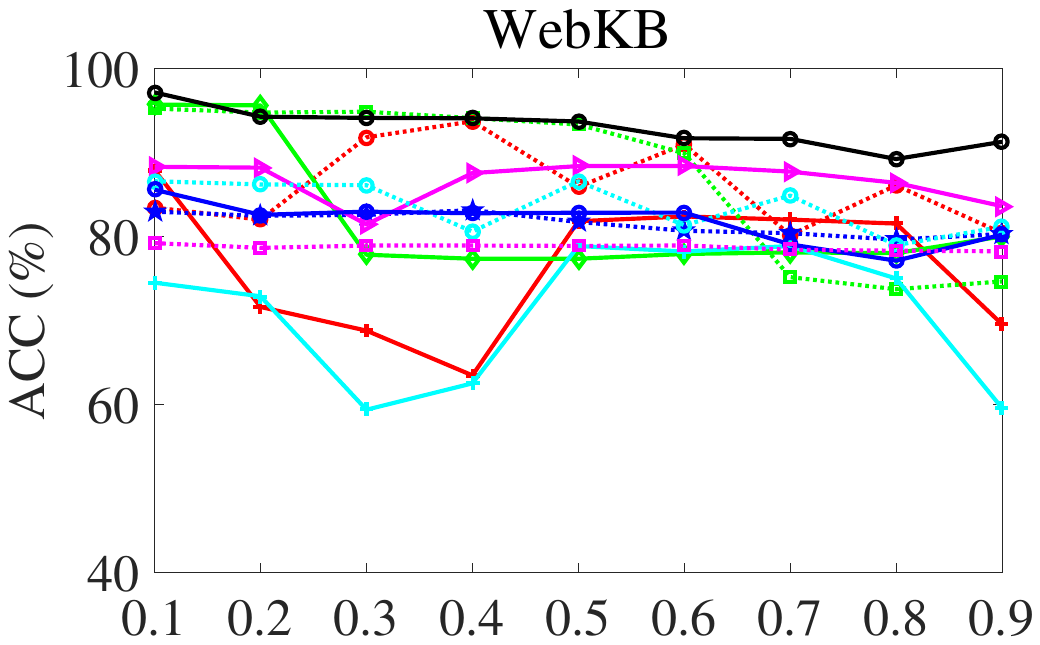}
}
\subfigure{
\includegraphics[trim=50 250 60 280, width=0.235\textwidth]{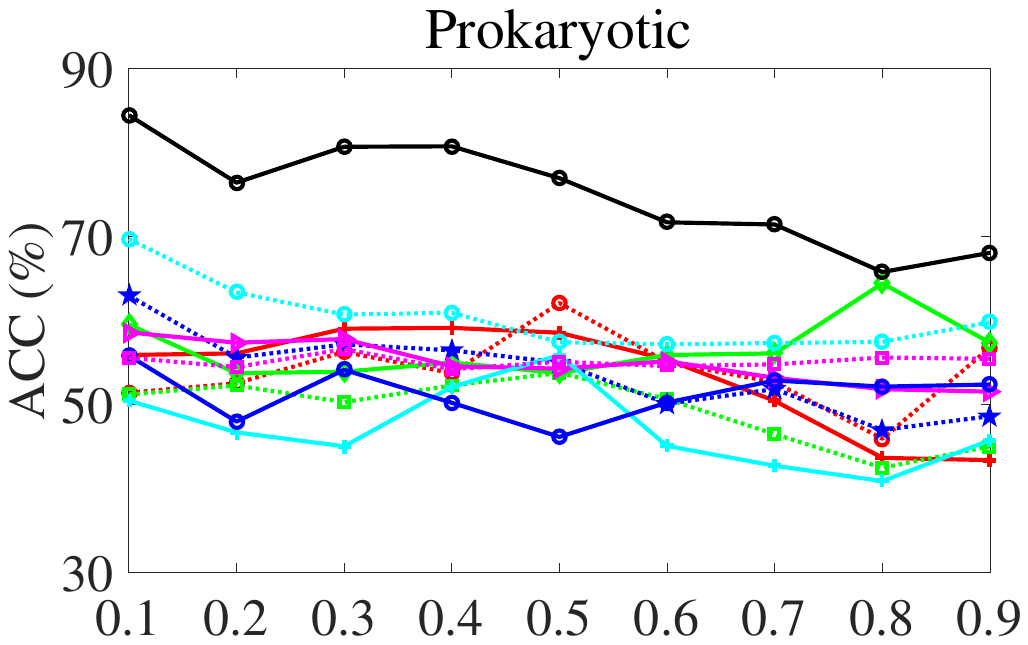}
}
\subfigure{
\includegraphics[trim=50 250 60 280, width=0.235\textwidth]{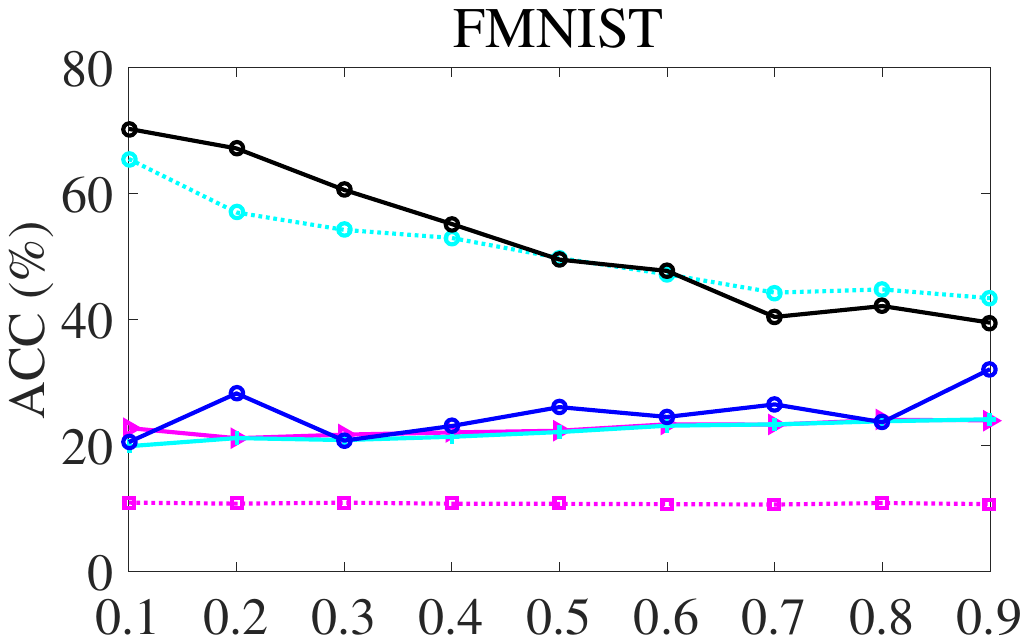}
}
% \subfigure{
% \includegraphics[width=0.23\textwidth]{ACC_CIFAR10.pdf}
% }
\subfigure{
\includegraphics[trim=50 250 60 280, width=0.235\textwidth]{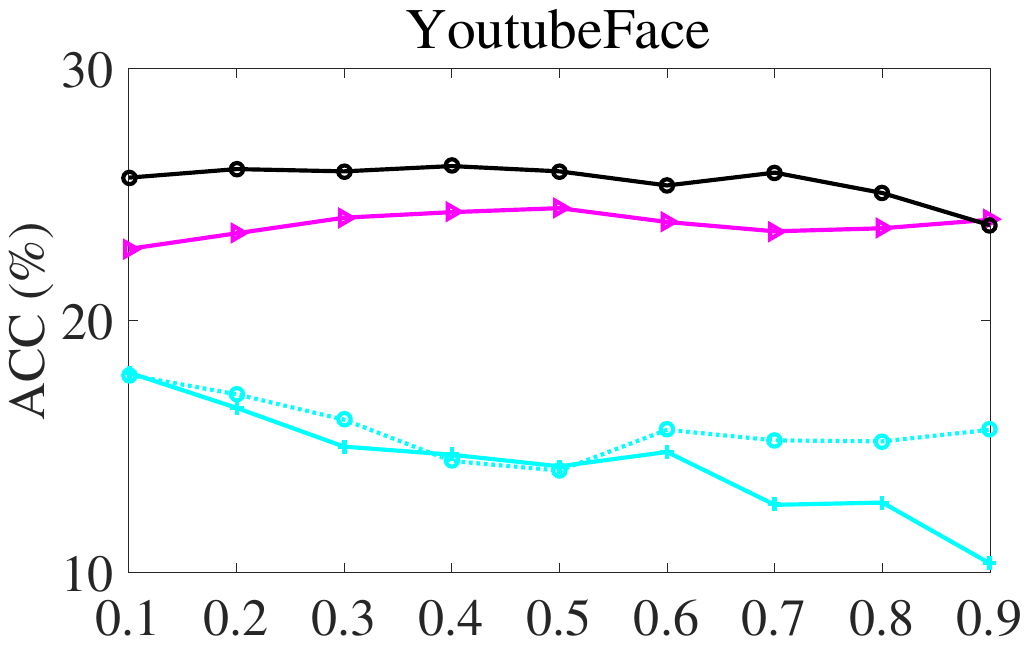}
}
\vspace{-10pt}

\subfigure{
\includegraphics[trim=50 250 60 280, width=0.235\textwidth]{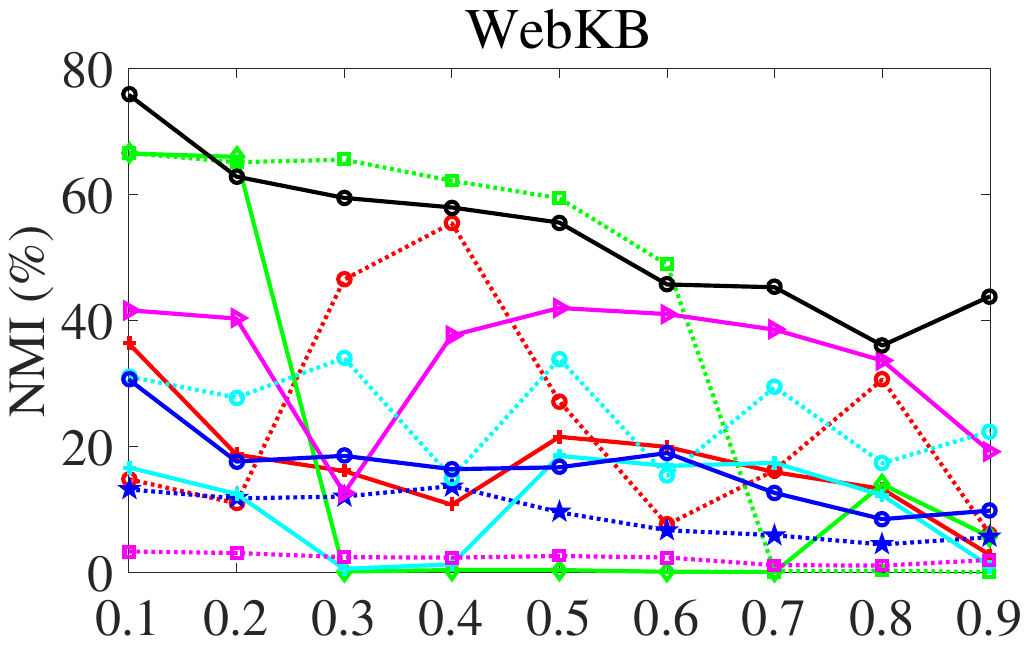}
}
\subfigure{
\includegraphics[trim=50 250 60 280, width=0.235\textwidth]{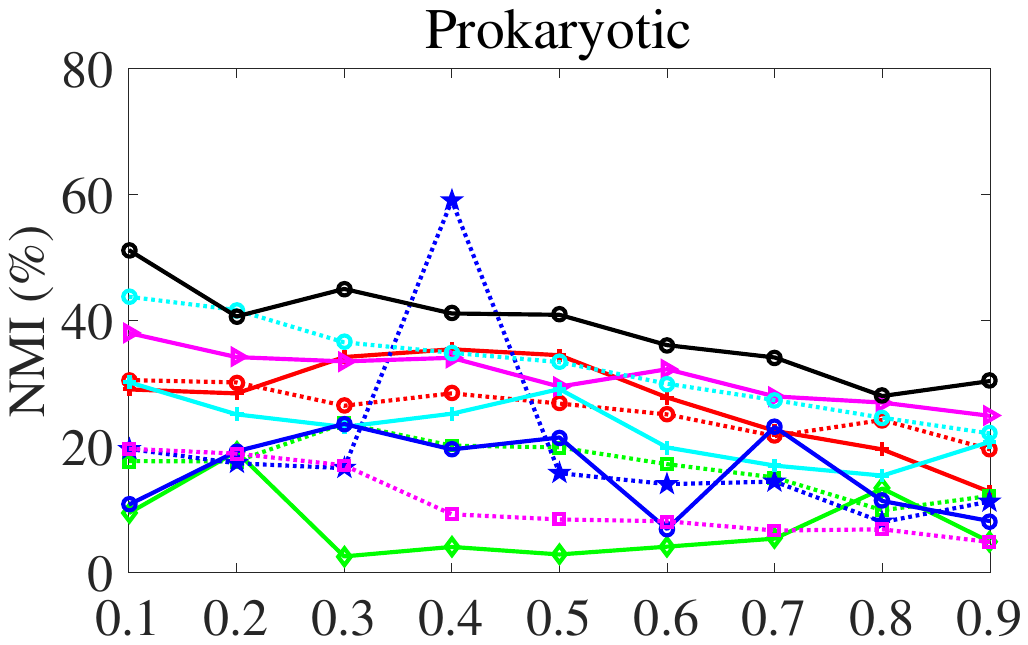}
}
\subfigure{
\includegraphics[trim=50 250 60 280, width=0.235\textwidth]{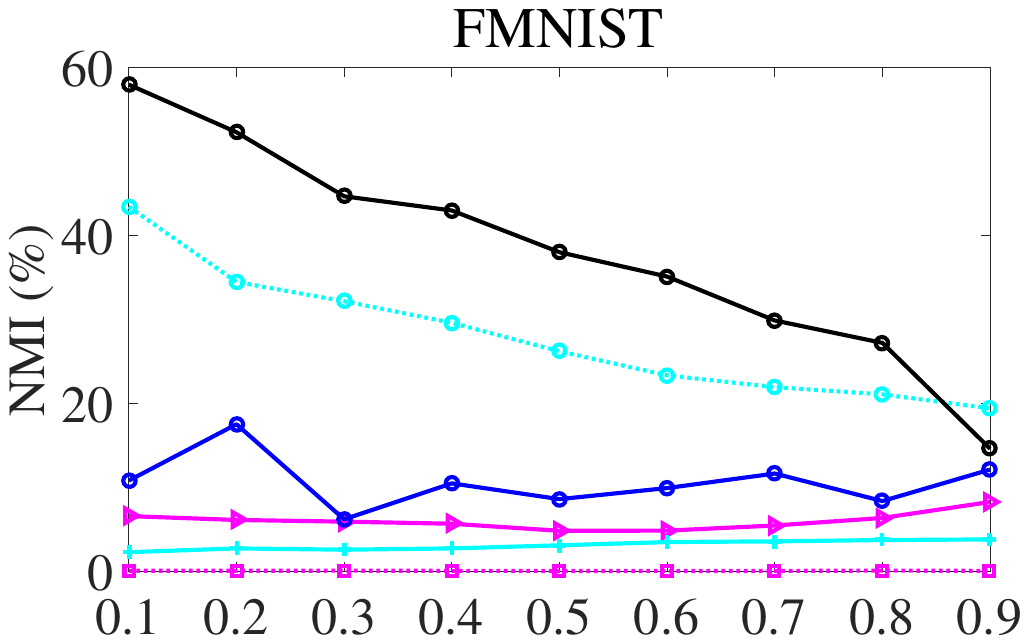}
}
% \subfigure{
% \includegraphics[width=0.23\textwidth]{NMI_CIFAR10.pdf}
% }
\subfigure{
\includegraphics[trim=50 250 60 280, width=0.235\textwidth]{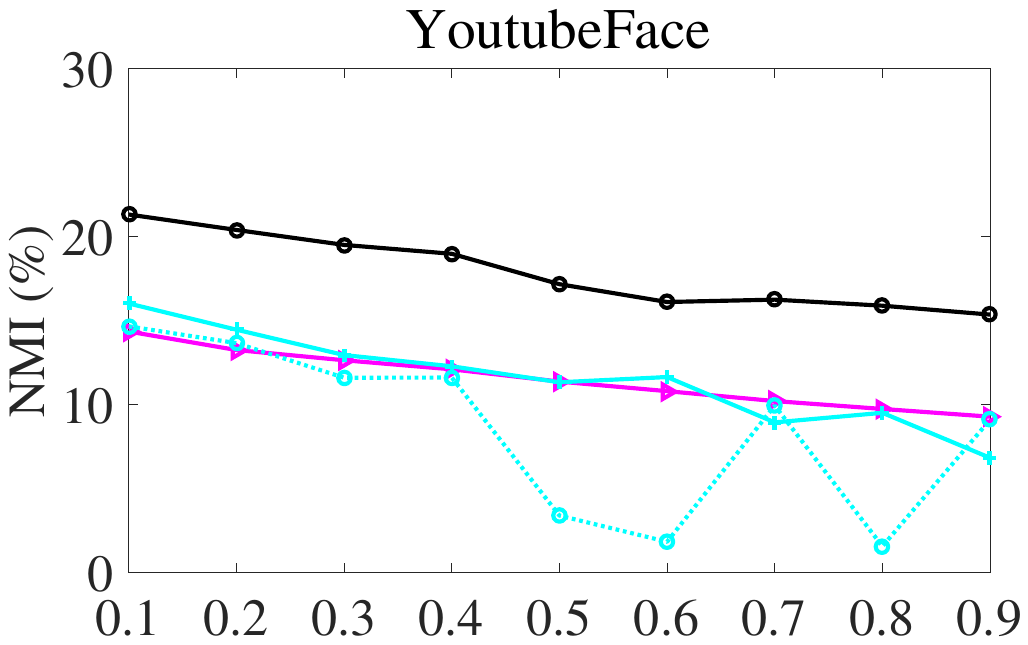}
}
\vspace{-10pt}

\subfigure{
\includegraphics[trim=37 250 60 270, width=0.235\textwidth]{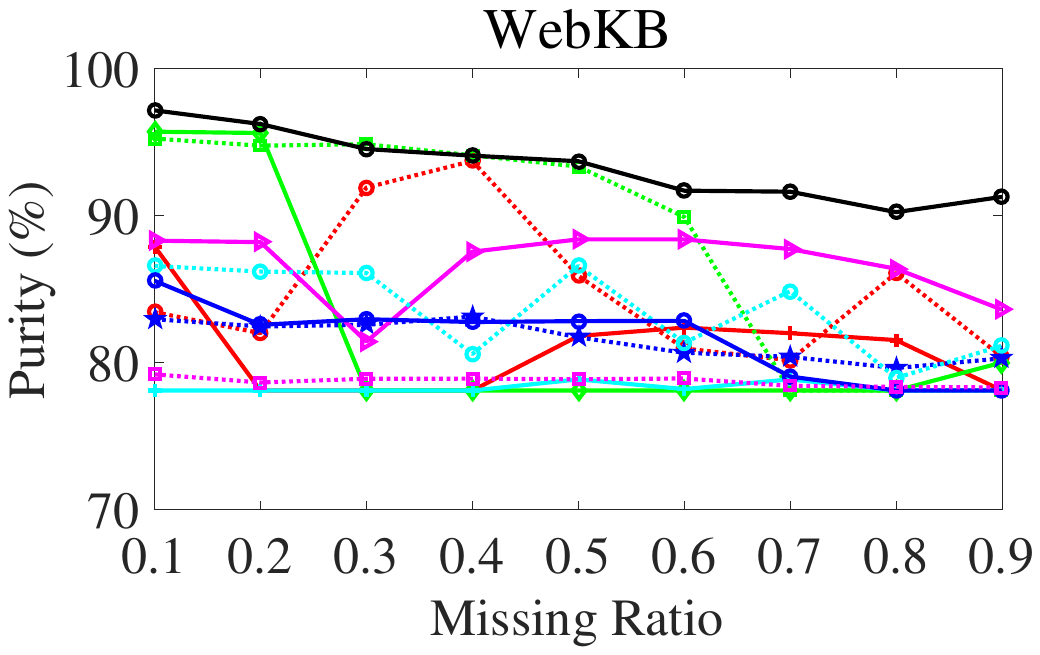}
}
\subfigure{
\includegraphics[trim=50 250 60 270, width=0.235\textwidth]{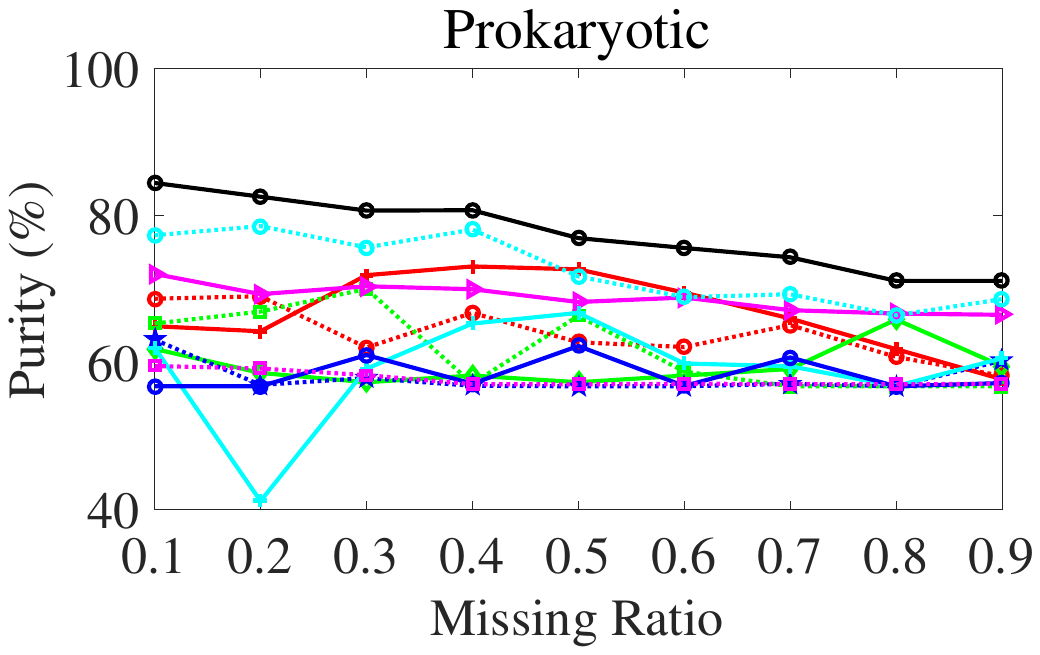}
}
\subfigure{
\includegraphics[trim=50 250 60 270, width=0.235\textwidth]{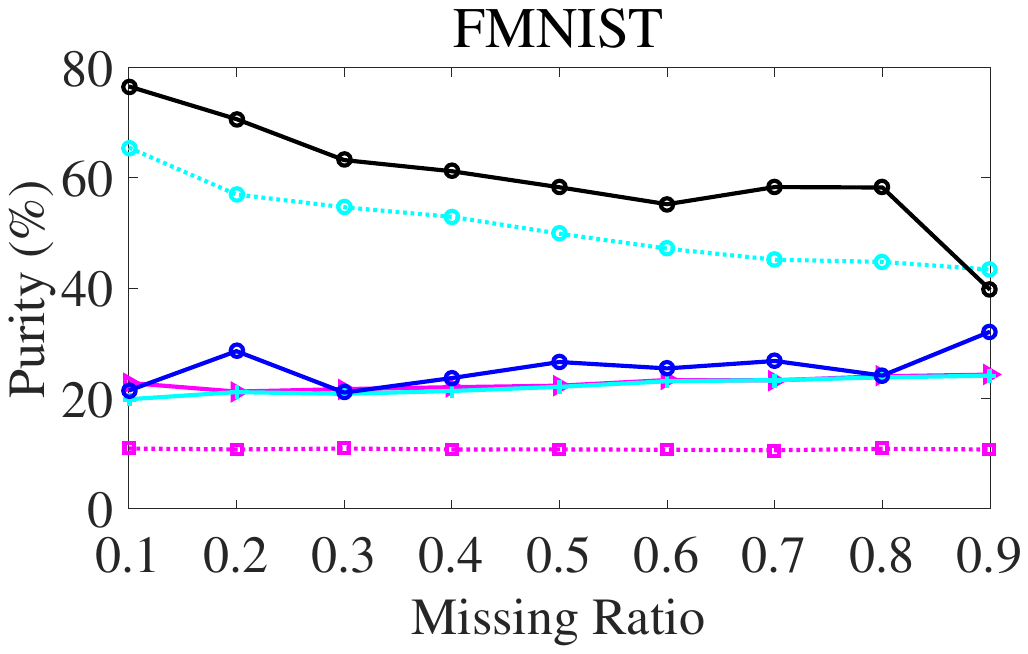}
}
% \subfigure{
% \includegraphics[width=0.23\textwidth]{Purity_CIFAR10_2.pdf}
% }
\subfigure{
\includegraphics[trim=50 250 60 270, width=0.235\textwidth]{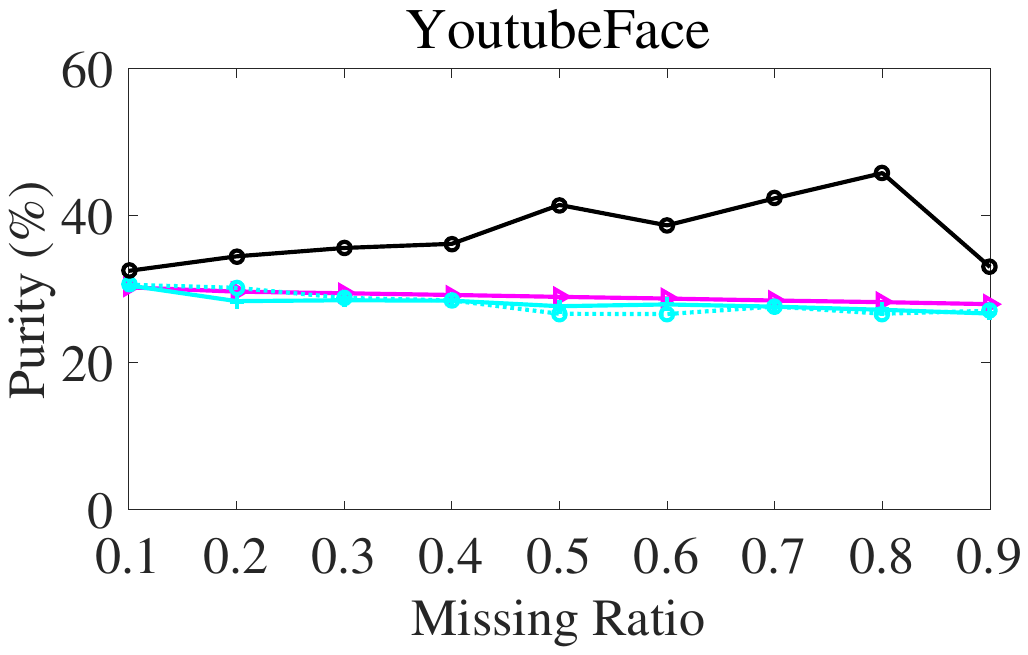}
}
\vspace{-8pt}

\subfigure{
\includegraphics[width=0.5 \textwidth]{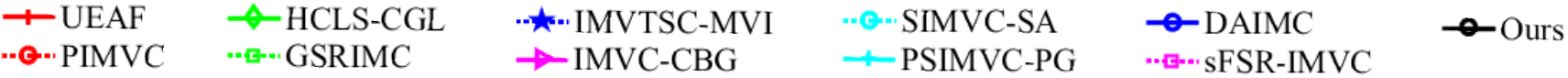}
}
\vspace{-5pt}
\caption{The ACC, NMI and Purity results of different methods with different missing ratios on partial benchmark datasets.}
\label{fig:performance}
\end{figure*}

\begin{table*}[ht]
% \footnotesize
\centering
\scriptsize
\renewcommand\arraystretch{0.9}
\tabcolsep = 3.5 pt
\begin{tabular}{c|cccccccccccc}
\hline
Datasets     & UEAF    & PIMVC         & HCLS-CGL & GSRIMC  & IMVTSC-MVI & DAIMC    & sFSR-IMVC & PSIMVC-PG & IMVC-CBG & SIMVC-SA & Ours        \\ \hline
Prokaryotic  & 99.15   & 2.94          & 26.72    & 128.92  & 352.19     & 73.45    & 7.72      & 8.71      & 17.25    & 16.74    & \textbf{0.37}   \\
WebKB        & 526.35  & 7.20          & 21.14    & 239.32  & 1024.35    & 651.86   & 10.31     & 13.27     & 17.08    & 11.49    & \textbf{0.66}   \\
Caltech101-7 & 635.6   & \textbf{9.82} & 165.91   & 623.22  & 8433.70    & 1120.39  & 15.15     & 28.41     & 37.87    & 50.29    & {\underline{11.63}}     \\
Wikipedia    & 2344.69 & 9.37          & 280.86   & 2366.09 & 3916.00    & 83.10    & 27.01     & 28.33     & 28.92    & 9.21     & \textbf{2.84}   \\
CIFAR10      & -       & -             & -        & -       & -          & 93468.10 & 520.11    & 596.21    & 627.45   & 1359.07  & \textbf{149.99} \\
FMNIST       & -       & -             & -        & -       & -          & 89872.00 & 1602.51   & 846.43    & 567.30   & 973.02   & \textbf{371.89} \\
YoutubeFace  & -       & -             & -        & -       & -          & -        &    -       & 1423.55   & 1866.46  & 3209.01  & \textbf{899.54} \\ \hline
\end{tabular}
\caption{Time costs of different methods on benchmark datasets (measured in seconds). ``-'' means out of CPU memory. The best results are highlighted in bold, while the second-best results are marked with underline.}
\label{times}
\end{table*}

\noindent\textbf{Compared Methods.}
We compare our RISE with ten state-of-the-art IMVC methods. These methods include UEAF \cite{wenjie2019unified}, PIMVC \cite{deng2023projective}, HCLS-CGL \cite{wenjie2023highly}, GSRIMC \cite{li2022refining}, IMVTSC-MVI \cite{wenjie2021unified}, sFSR-IMVC\cite{long2024feature}, 
DAIMC \cite{hu2019doubly}, PSIMVC-PG \cite{li2024parameter}, IMVC-CBG \cite{wangsiwei2022highly}, and SIMVC-SA \cite{wenyi2023scalable}. 

\noindent\textbf{Implementation Details.}
For our RISE method, we use $K$-means for anchor selection and the approach in \cite{hefang2020fast} to construct the initial bipartite graph $\{B^{(i)}\}_{i=1}^v$. We tune the parameter $\beta$ within a range of $[0.01,0.1,1,10,20,50,100,500,1000]$. For the number of anchor points, $m$ is tuned within a range of $[1c,6c]$ for the Prokaryotic, CIFAR10, FMNIST, and YoutubeFace datasets, and $[2c,10c]$ for the Wikipedia and WebKb datasets. Moreover, the embedded dimension $k$ is adjusted as an integer multiple of the number of categories and kept smaller than the number of anchors. For all competitors, we obtain their public source codes from open websites and run these algorithms with their default parameter settings. Otherwise, we perform parameter search for better performance. To alleviate the sensitivity of random initialization, we repeat all experiments 10 times and report the average performance. Following previous works \cite{zhang2019multi}, clustering accuracy (ACC), normalized mutual information (NMI), and purity are utilized as metrics to evaluate the performance of all methods. All experiments are performed on a machine with Inter core i7-9700 CPU, 32GB RAM, and Matlab 2022b (64bit).

\begin{figure}[!t]
% \centering
% \setlength{\abovecaptionskip}{0cm}
\subfigure{
\includegraphics[trim=60 290 120 300, width=0.224\textwidth]{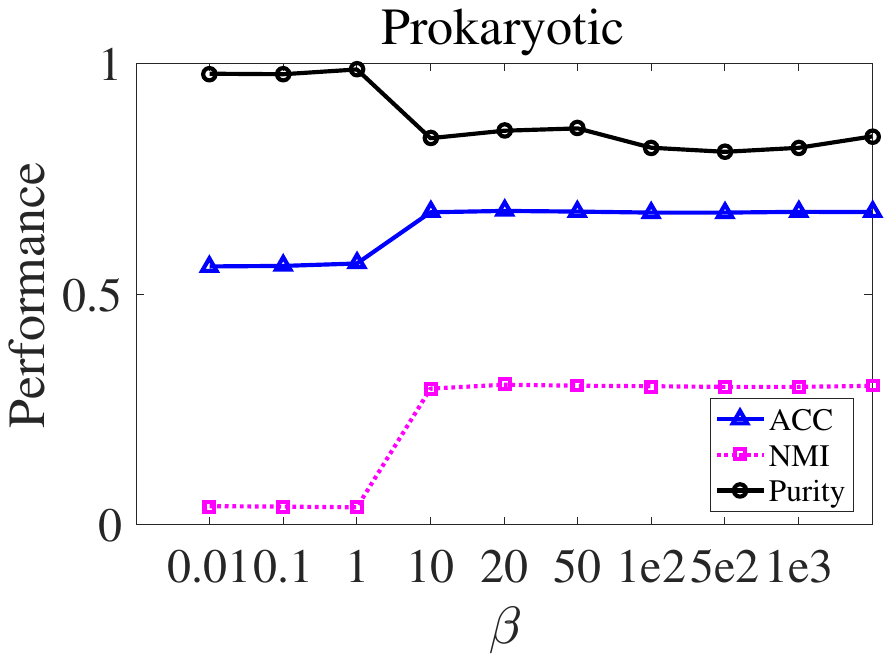}
}
\subfigure{
\includegraphics[trim=60 290 120 300, width=0.224\textwidth]{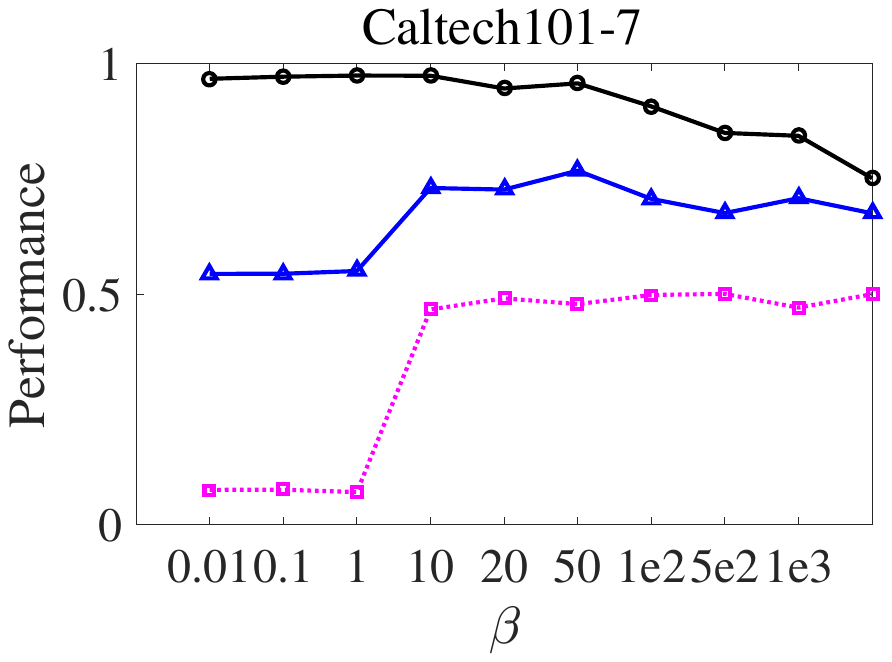}
}
\caption{Clustering results of our RISE with different values of $\beta$ on two datasets.}
\label{fig:sensitivity}
\end{figure}

\begin{figure}[!t]
\centering
\subfigure{
\includegraphics[trim=130 268 140 270, clip, width=0.224\textwidth]{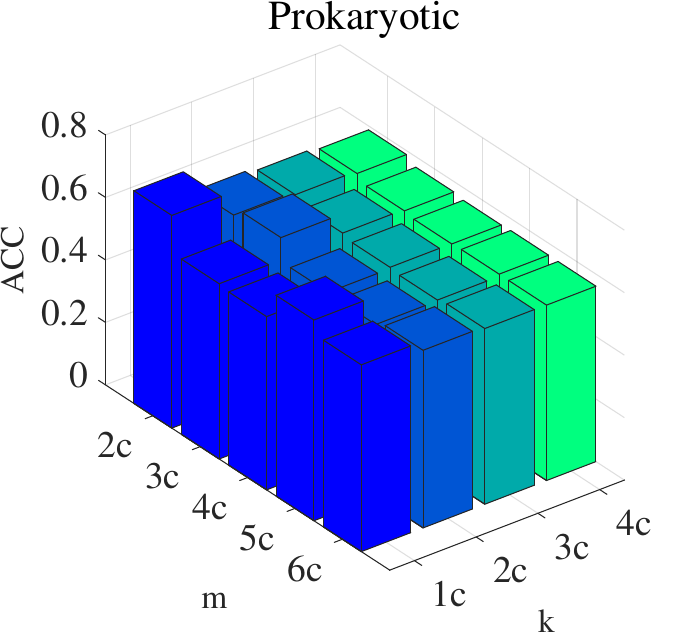}
}
\subfigure{
\includegraphics[trim=130 268 140 270,clip, width=0.224\textwidth]{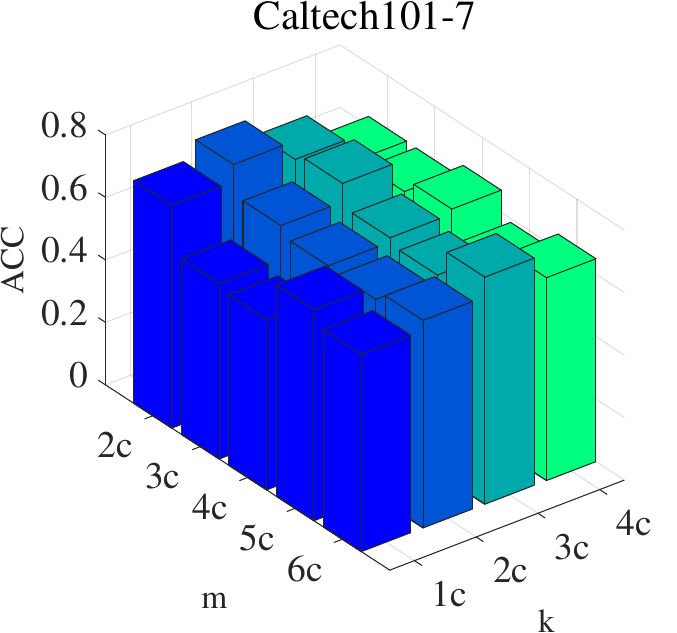}
}
\caption{The ACC results of our RISE with different values of $m$ and $k$ on two datasets.}
\label{fig:anchor2dim}
\end{figure}

\subsection{Experimental Results and Analysis}
Table \ref{table_performance} presents the average clustering results of different methods with nine missing ratios. We can draw the following conclusions.

\begin{enumerate} 
  \item Our proposed RISE achieves the best performance across the three metrics against all baseline algorithms in most cases, indicating its superiority in IMVC tasks. Notably, SIMVC-SA performs better than other methods. Our RISE surpasses SIMVC-SA with improvements of 14.71\%, 9.42\%, 15.88\%, 1.42\%, 0.5\%, 1.5\%, and 9.84\% in terms of ACC on the Prokaryotic, WebKB, Caltech101-7, Wikipedia, CIFAR10, FMNIST, and YoutubeFace datasets, respectively. Similar trends are observed across other metrics, further demonstrating the effectiveness of our RISE, even on large-scale datasets.
  \item Compared to instance-level methods (i.e., UEAF, PIMVC, IMVTSC-MVI, sFSR-IMVC, and DAIMC) and graph-level methods (i.e., HCLS-CGL, GSRIMC, PSIMVC-PG, IMVC-CBG, and SIMVC-SA), our RISE based on spectral embedding completion demonstrates its effectiveness. 
\end{enumerate} 

To provide a more intuitive comparison of different methods across various missing ratios, we present the ACC, NMI, and Purity curves for partial datasets in Fig. \ref{fig:performance}. As shown in Fig. \ref{fig:performance}, our RISE outperforms other methods, demonstrating superior and more stable performance. This highlights the advantage of the second-order rotational-invariant property in multi-view fusion.

Table \ref{times} presents the running times consumed by different methods. As illustrated, our RISE requires the shortest running time on all datasets with the exception of Caltech101-7. This demonstrates the efficiency of RISE. Note that SIMVC-SA and IMVC-CBG take 3209.01 seconds and 1866.46 seconds on the YoutubeFace dataset, while our method only takes 899.54 seconds, reducing the running time by over half. This is because RISE employs bipartite graphs to preserve the essential intrinsic geometric structure of data while reducing redundancy. Moreover, the proposed optimization algorithm further accelerates the clustering process.

\begin{figure}[!t]
\vspace{-10pt}
\centering
\setlength{\abovecaptionskip}{0cm}
\subfigure{
\includegraphics[trim=92 245 118 270,clip, width=0.224\textwidth]{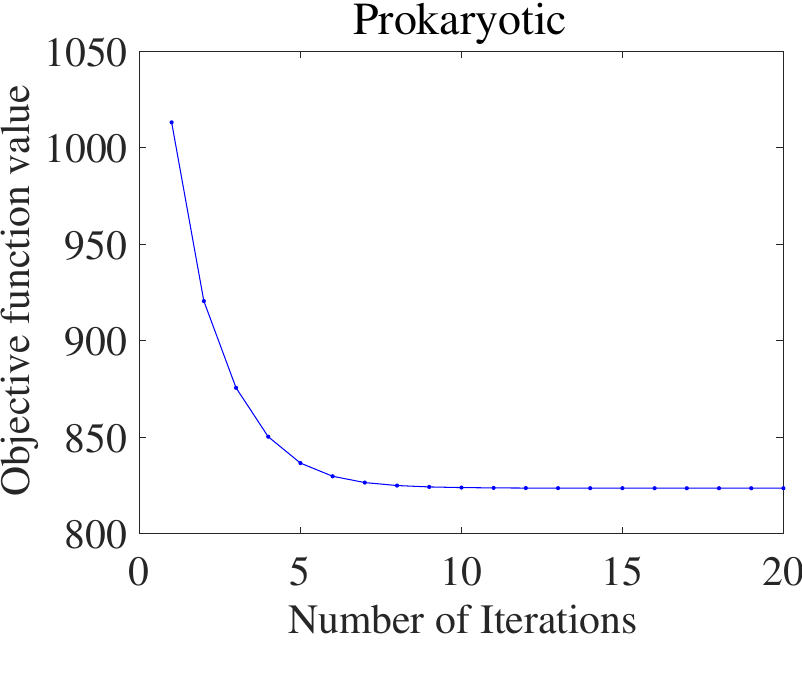}
}
\subfigure{
\includegraphics[trim=92 245 118 270,clip, width=0.224\textwidth]{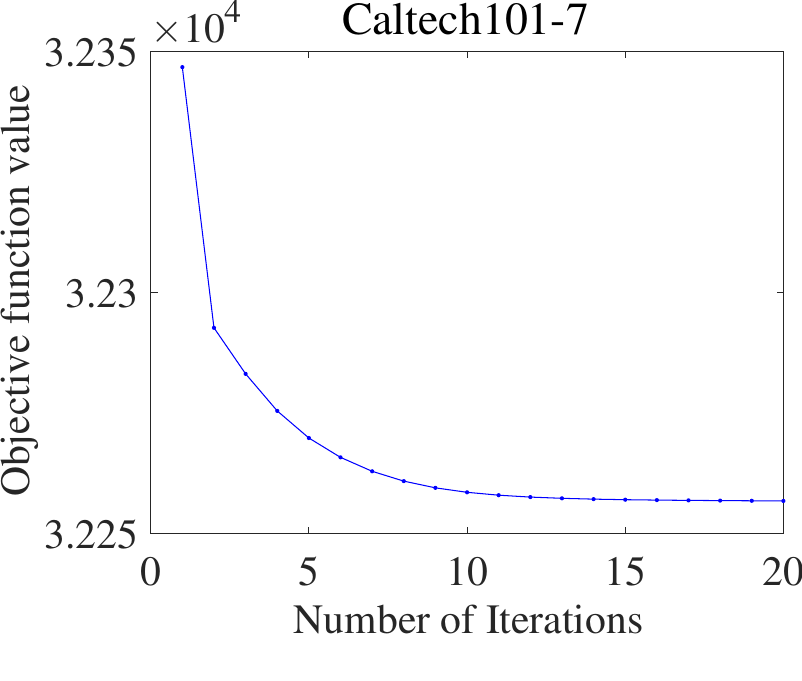}
}
\caption{The variation of the objective function values for our RISE on two datasets.}
\label{fig:convergence}
\end{figure}

\subsection{Parameter Sensitivity and Convergence Analysis}
To investigate the parameter sensitivity of RISE, we conducted experiments with different parameter settings on two datasets. As shown in Fig. \ref{fig:sensitivity}, the parameter $\beta$ indeed affects performance. Specifically, the proposed algorithm remains stable over a large range of $\beta$, except when it is less than $10$, in which case the NMI metric suffers from a significant degradation. 
Additionally, we investigated the effect of varying the number of anchors $m$ and embedded dimensions $k$ under the optimal setting of other variables. As shown in Fig \ref{fig:anchor2dim}, both the number of anchors and embedded dimension influence the performance of our algorithm. This means that the appropriate values for $m$ and $k$ need to be chosen for RISE to obtain the optimal results. Fortunately, only a small amount of tuning is required to achieve satisfactory results.

In addition, we conducted several experiments to validate the convergence of the proposed algorithm. As shown in Fig. \ref{fig:convergence}, the objective function value of our algorithm decreases monotonically in each iteration and converges rapidly. These results verify the convergence of our proposed algorithm.

\subsection{Ablation Study}
\begin{figure}[!t]
\centering
\subfigure{
\includegraphics[trim=85 265 130 280, width=0.224\textwidth]{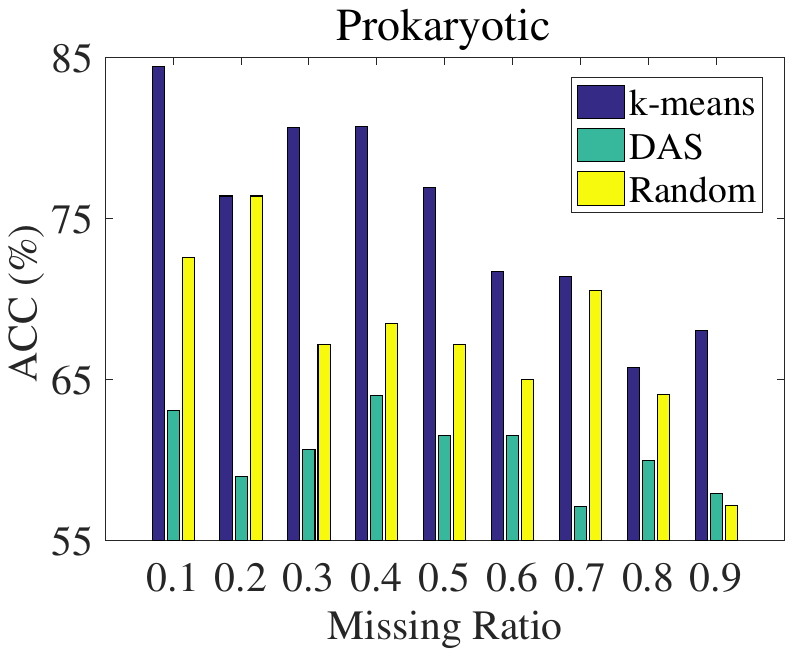}
}
\subfigure{
\includegraphics[trim=85 265 130 280, width=0.224\textwidth]{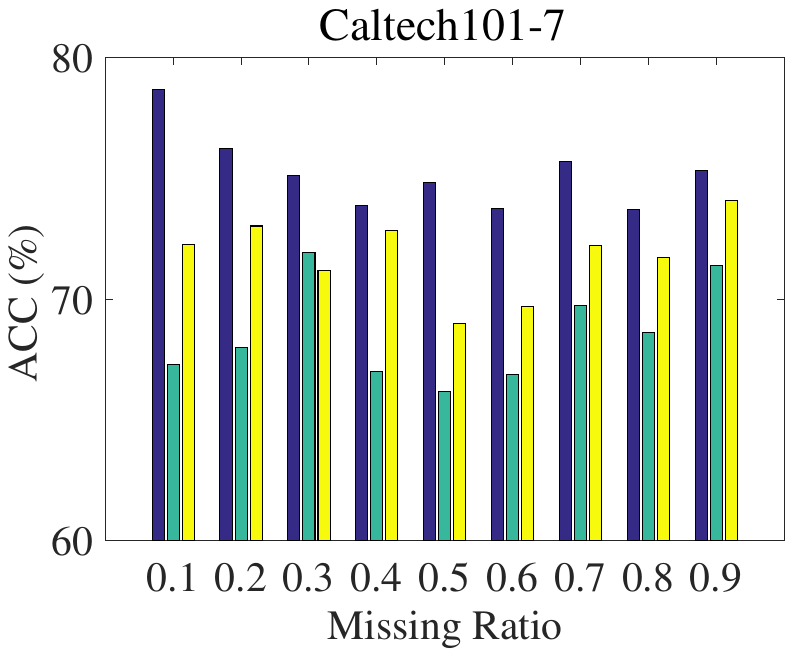}
}
\vspace{-10pt}
\caption{The ACC results of our RISE with different anchor selection strategies on two datasets.}
\label{fig:anchor}
\end{figure}

\textbf{Effect of Anchor Selection Strategy.}
To investigate the effect of different strategies for generating anchors, we used the k-means, random selection, and directly alternate sampling (DAS) method \cite{lixuelong2020multiview} for anchor selection. As shown in Fig. \ref{fig:anchor}, the k-means is more suitable for our algorithm. Specifically, the $k$-means based anchors achieve over 15\% performance improvement compared to the DAS approach on the Prokaryotic dataset, and around 10\% improvement on the Caltech101-7 dataset. This is likely because the DAS approach is designed for complete multi-view data, while $k$-means is a more general method.

\begin{figure}[!t]
\centering
\setlength{\abovecaptionskip}{0cm}
\subfigure{
\includegraphics[trim=75 250 125 270,clip, width=0.224\textwidth]{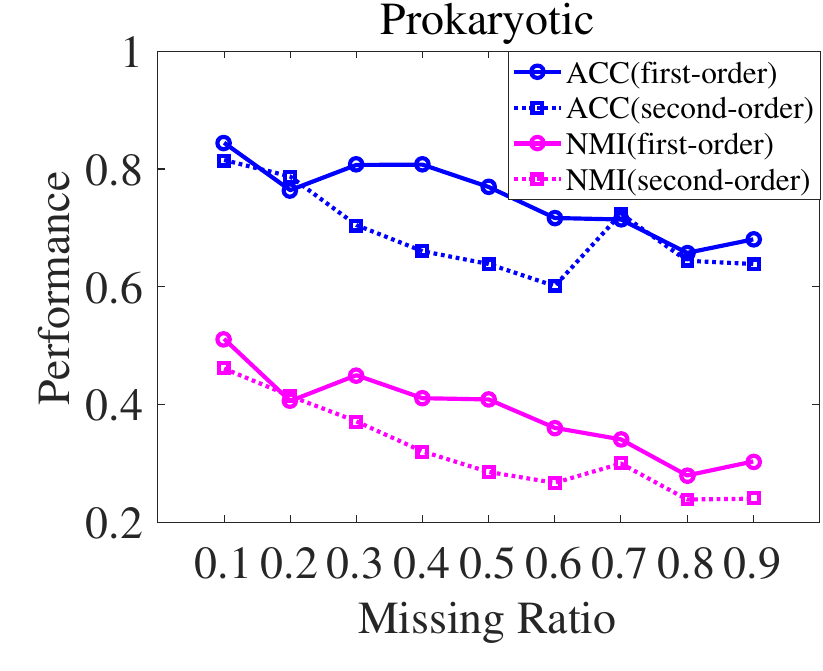}
}
% \subfigure{
% \includegraphics[width=0.22\textwidth]{strategy_WebKB1.pdf}
% }
\subfigure{
\includegraphics[trim=75 250 125 270,clip, width=0.224\textwidth]{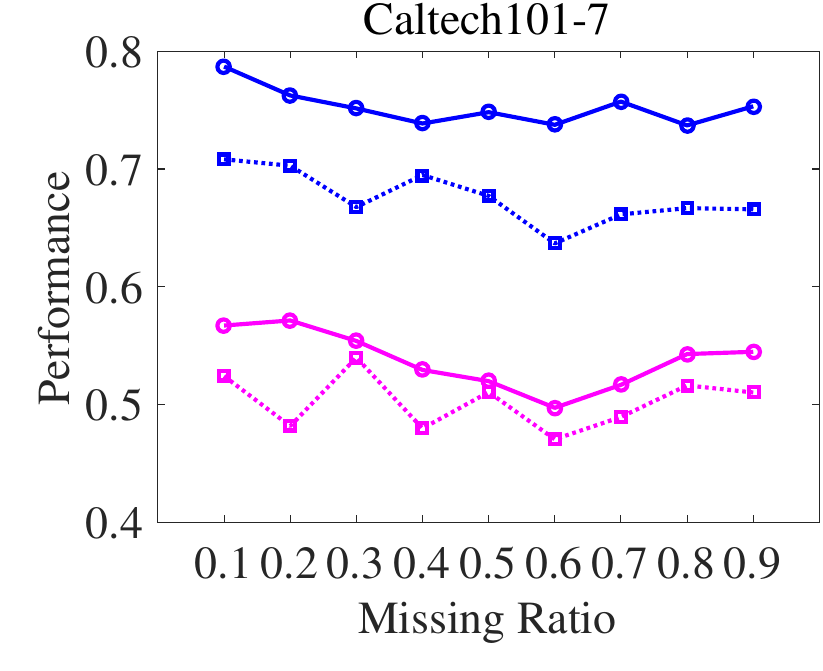}
}
\caption{Clustering results of our RISE with different completion strategies on two datasets.}
\label{fig:ablation}
\end{figure}

\noindent\textbf{Performance Effect of Completion Strategy.}
The rotation-invariant complete representation learning strategy is the main contribution of this paper. To further demonstrate its effectiveness, we conducted ablation experiments comparing different complete representation learning strategies. The term 'second-order' refers to our proposed method, while the term 'first-order' indicates the use of Eq. \eqref{eq:3-1-1} to learn the complete representation. As shown in Fig. \ref{fig:ablation}, the proposed rotation-invariant learning module effectively improves the algorithm's performance on the Prokaryotic and WebKB datasets in terms of ACC and NMI. This demonstrates the effectiveness of our proposed strategy. Additionally, our strategy shows smaller performance fluctuations as the missing rate increases, indicating the superiority of our approach. 

\begin{figure}[!t]
\centering
\setlength{\abovecaptionskip}{0cm}
\subfigure{
\includegraphics[trim=85 250 130 280, width=0.224\textwidth]{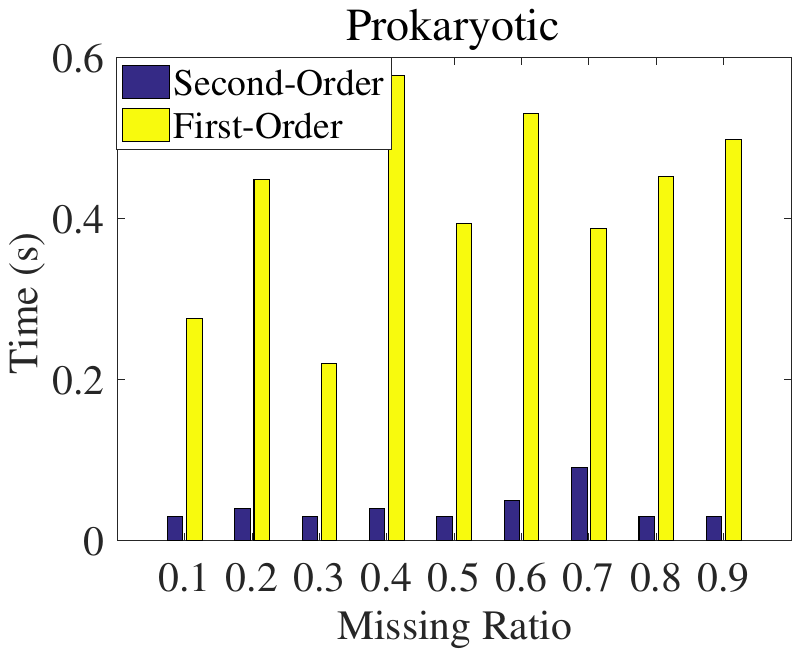}
}
\subfigure{
\includegraphics[trim=85 250 130 280, width=0.224\textwidth]{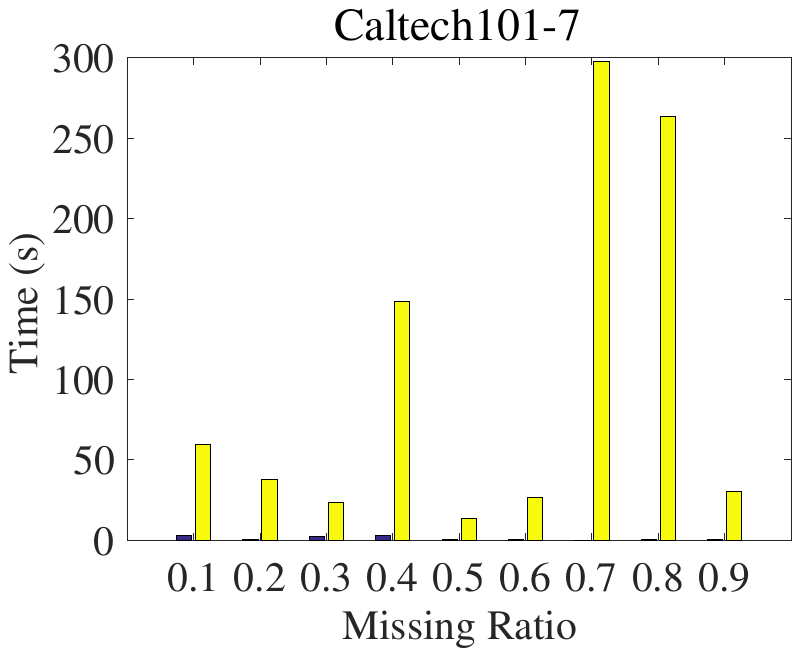}
}
\caption{Time costs of our RISE with 
different completion strategies on two datasets.}
\label{fig:ablation_times1}
\end{figure}

\begin{figure}[!t]
\centering
\setlength{\abovecaptionskip}{0cm}
\subfigure{
\includegraphics[trim=75 250 125 270,clip, width=0.224\textwidth]{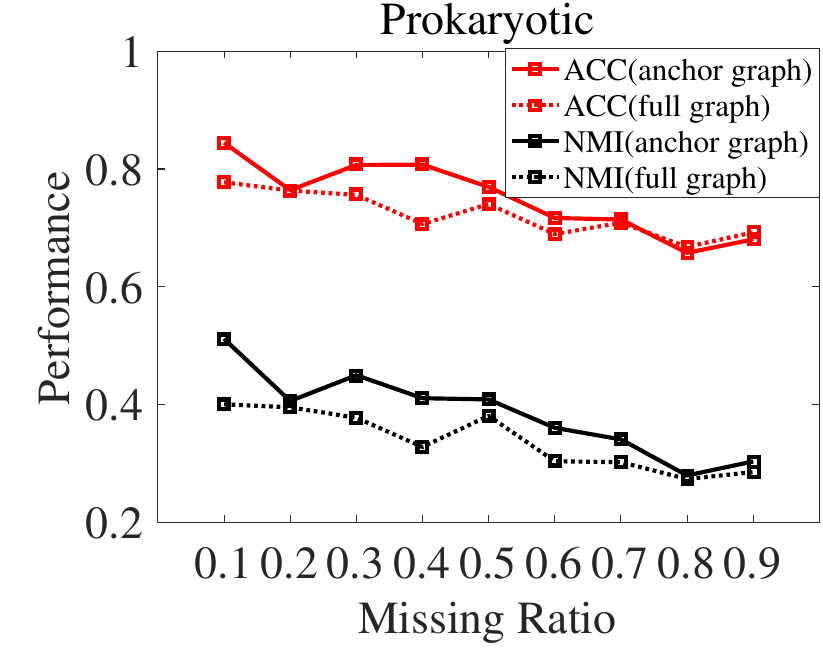}
}
\subfigure{
\includegraphics[trim=75 250 125 270,clip, width=0.224\textwidth]{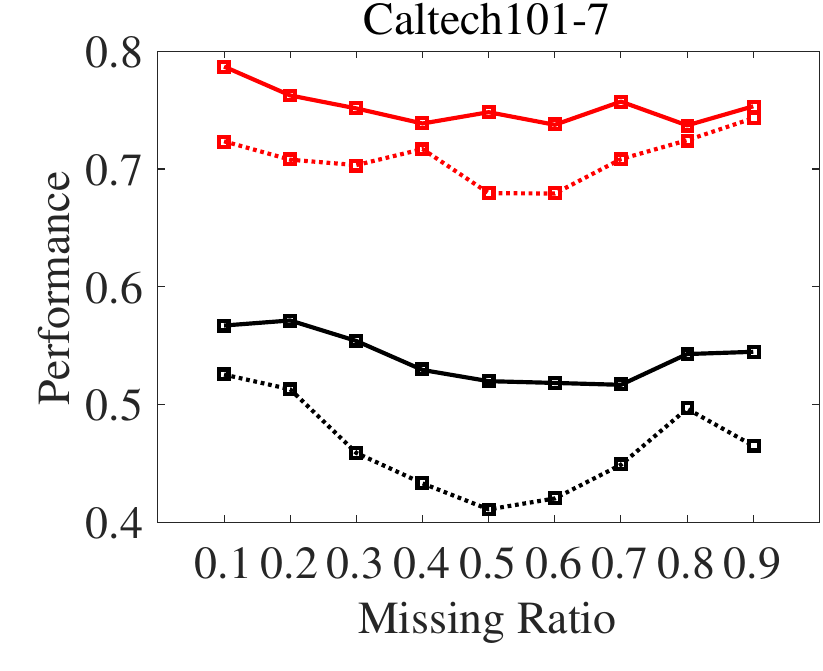}
}
\caption{Clustering results with bipartite graph and full-size similarity graph on two datasets.}
\label{fig:ablation_anchor2full}
\end{figure}

\noindent\textbf{Efficiency Effect of Completion Strategy.}
The proposed optimization algorithm enhances the efficiency of our RISE method. To further demonstrate its effectiveness, we plot the execution time comparison histograms of learning strategies using Eq. \eqref{eq:3-1-1} and our proposed second-order rotation-invariant complete representation learning in Fig. \ref{fig:ablation_times1}. From this, we observe that our RISE method requires less time compared to the first-order learning strategy using Eq. \eqref{eq:3-1-1}. Specifically, our RISE shows nearly 10 times lower execution time on the Prokaryotic dataset and around 100 times lower on the Caltech101-7 dataset.

\noindent\textbf{Effect of Graph Selection.}
To evaluate the performance of bipartite graph in reducing data redundancy, we used both bipartite graph and full similarity graph to separately describe the neighborhood structure among data. We then executed the rotation-invariant spectral embedding framework to obtain the consensus complete representation and applied $K$-means to get the final clustering indicators. As is shown in Fig. \ref{fig:ablation_anchor2full}, the bipartite graph can provides more superior performance compared to the full similarity graph. Additionally, using full similarity graph results in higher optimization costs because the proposed simplified SVD-based optimization strategy cannot be used.

\section{Conclusions}
This paper proposed a highly efficient incomplete multi-view learning framework, called RISE. It developed a fast late spectral embedding completion technique. RISE learned view-specific embeddings from incomplete bipartite graphs and then recovered the complete consensus representation from these embeddings using the proposed second-order rotation-invariant learning module. Both processes were integrated into a unified framework to mutually enhance performance. To solve the proposed objective function, we designed a fast two-step alternating optimization algorithm based on double SVD decompositions. Extensive experiments on datasets of varying sizes demonstrated the effectiveness, scalability, and efficiency of our proposed method. In the future, developing a parameter-free model will require more attention.

% \clearpage
% \appendix

% \section{Appendices}
% \subsection{Structure of this document}
% This document is organized as follows. Firstly, it provides some preliminaries for multi-view spectral clustering and bipartite graph learning. Secondly, it reports the detailed experimental configurations. Thirdly, it presents the proofs of convergence of the proposed optimization algorithm. 

\section{Acknowledgments}
This work was funded by the Science and Technology Development Fund, Macau SAR (File no. 0049/2022/A1, 0050/2024/AGJ), by the University of Macau Development Foundation (File no. MYRG-GRG2024-00181-FST-UMDF). 

\bibliography{aaai25}

\end{document}